\documentclass[10pt,twocolumn,letterpaper]{article}

\usepackage[pagenumbers]{cvpr} 

\makeatletter
\@namedef{ver@everyshi.sty}{}
\makeatother
\usepackage{tikz}
\usepackage{graphicx}
\usepackage{amsmath}
\usepackage{amssymb}
\usepackage{booktabs}
\usepackage{csquotes}
\usepackage{colortbl}

\usepackage{booktabs}

\usepackage{array} 
\newcolumntype{M}[1]{>{\centering\arraybackslash}m{#1}}

\newcommand{\darkleftbox}[3]{%
\begin{tikzpicture}\node[inner sep=0pt] (p00) {\includegraphics[width=#1]{#2}};\node[rectangle, inner sep=1pt, above right, fill=black, fill opacity=0.6, text opacity=1.] at (p00.south west) {\footnotesize \textcolor{white}{#3}}; \end{tikzpicture}%
}
\newcommand{\darkleftrightbox}[4]{%
\begin{tikzpicture}
\node[inner sep=0pt] (p00) {\includegraphics[width=#1]{#2}};
\node[rectangle, inner sep=1pt, above right, fill=black, fill opacity=0.6, text opacity=1.] at (p00.south west) {\footnotesize \textcolor{white}{#3}};
\node[rectangle, inner sep=1pt, above right, fill=black, fill opacity=0.6, text opacity=1., anchor=north east] at (p00.north east) {\footnotesize \textcolor{white}{#4}};
\end{tikzpicture}%
}
\newcommand{\lightrightbox}[3]{%
\begin{tikzpicture}\node[inner sep=0pt] (p00) {\includegraphics[width=#1]{#2}};\node[rectangle, inner sep=1pt, above right, fill=white, fill opacity=0.6, text opacity=1., anchor=north east] at (p00.north east) {\footnotesize \textcolor{black}{#3}}; \end{tikzpicture}%
}

\newcommand{\targ}{f^T}
\newcommand{\fadv}{\check{f}}

\newcommand{\loss}{\mathcal{L}}

\newcommand{\de}{\delta_{t}}
\newcommand{\dt}{\delta_{t\!+\!1}}
\newcommand{\transp}{\theta}

\usepackage[pagebackref,breaklinks,colorlinks]{hyperref}

\usepackage[capitalize]{cleveref}
\crefname{section}{Sec.}{Secs.}
\Crefname{section}{Section}{Sections}
\Crefname{table}{Table}{Tables}
\crefname{table}{Tab.}{Tabs.}

\def\cvprWorkshopName{ECCV 2022 Workshop on Adversarial Robustness in the Real World, Tel Aviv, Israel.}

\begin{document}

\title{Attacking Motion Estimation with Adversarial Snow}

\author{Jenny Schmalfuss \hspace*{1.5cm} Lukas Mehl \hspace*{1.5cm} Andrés Bruhn\\
Institute for Visualization and Interactive Systems, University of Stuttgart, Germany\\
{\tt\small firstname.lastname@vis.uni-stuttgart.de}
}
\maketitle

\begin{abstract}
   Current adversarial attacks for motion estimation (optical flow) optimize small per-pixel perturbations, which are unlikely to appear in the real world.
   In contrast, we exploit a real-world weather phenomenon for a novel attack with adversarially optimized snow.
   At the core of our attack is a differentiable renderer that consistently integrates photorealistic snowflakes with realistic motion into the 3D scene.
   Through optimization we obtain adversarial snow that significantly impacts the optical flow while being indistinguishable from ordinary snow.
   Surprisingly, the impact of our novel attack is largest on methods that previously showed a high robustness to small L$_p$ perturbations.
\end{abstract}

\section{Introduction}
\label{sec:intro}

Adversarial attacks, which are a severe threat to neural networks, have recently been introduced in the context of optical flow.
There, the goal is to compute the pixel-wise 2D motion $f$ between two consecutive frames of an image sequence at times $t$ and $t\!+\!1$.
Current attacks \cite{Ranjan2019AttackingOpticalFlow, Schmalfuss2022PerturbationConstrainedAdversarial} modify these two frames in the 2D space and consequently ignore the actual 3D geometry of the scene and the objects moving within.
Moreover, when modifying pixels, they do not impose visual constraints, yielding attacked images that lack naturalism.
Therefore, the conclusions drawn from robustness analyses with these attacks might not necessarily reflect the robustness of optical flow methods in the real world -- where perturbations are more likely to appear in the form of weather phenomena.

\begin{figure}
\centering
\setlength{\fboxrule}{0.1pt}%
\setlength{\fboxsep}{0pt}%
\begin{tabular}{@{}M{27.2mm}@{\ }M{27.2mm}@{\ }M{27.2mm}@{}}
    \darkleftrightbox{27.2mm}{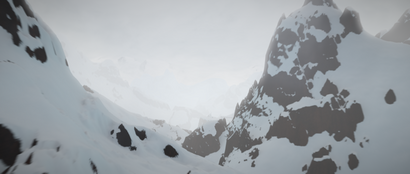}{Original}{Frame $t\vphantom{+}$} & \darkleftrightbox{27.2mm}{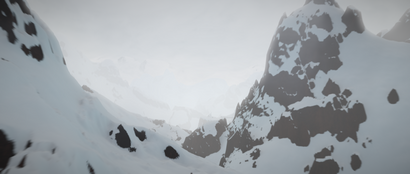}{}{Frame $t\!+\!1$} & \fcolorbox{gray!50}{white}{\lightrightbox
    {27.2mm}{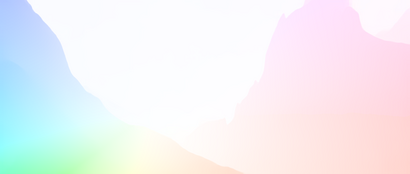}{Optical flow$\vphantom{+}$}} \\[-1pt]
    \darkleftbox{27.2mm}{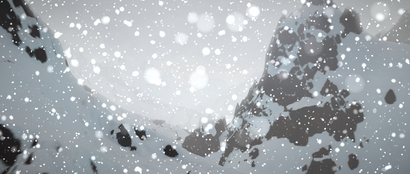}{Random snow} & \includegraphics[width=27.2mm]{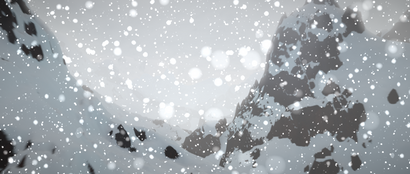} & \fcolorbox{gray!50}{white}{\includegraphics[width=27.2mm]{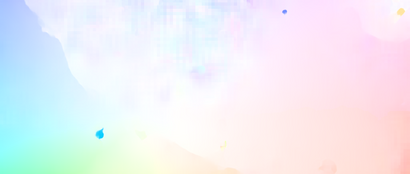}} \\[-1pt]
    \darkleftbox{27.2mm}{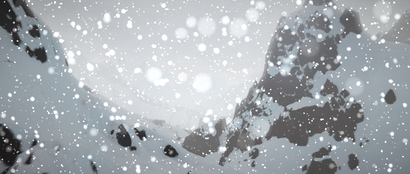}{Adversarial snow} & \includegraphics[width=27.2mm]{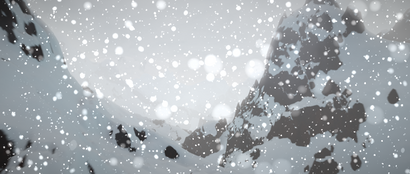} & \fcolorbox{gray!50}{white}{\includegraphics[width=27.2mm]{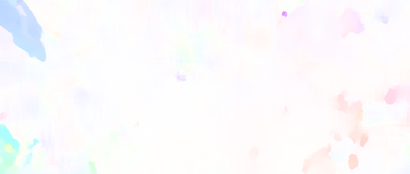}}\\
\end{tabular}
\caption{Snow attack with 3000 snowflakes that are first placed randomly in the 3D scene (\emph{Random snow}) and then optimized (\emph{Adversarial snow}) to perturb optical flow estimation with GMA~\cite{Jiang2021LearningEstimateHidden}.}
\label{fig:GMA attack}
\vspace{4mm}
\end{figure}

This work aims to answer the question whether a naturally occurring weather effect like snow can be manipulated to serve as an adversarial sample for motion estimation.
To this end, we propose an adversarial attack that augments images with falling snowflakes featuring a high degree of realism: 
We create snowflakes with a view-consistent 3D motion over time, insert them into the 3D scene in a depth-aware manner, and ensure photo-realism through visual effects (see Fig.~1).
This enables us to generate adversarially manipulated snow that significantly deteriorates optical flow predictions, while still satisfying the spatio-temporal and visual constraints of naturalistic snow.
We consider snow as representative weather effect where single particles move independent of the remaining scene content, but note that the proposed attack procedure could also be used to model rain or sleet.

\medskip \noindent
\textbf{Related work.}
Current optical flow methods based on neural networks\cite{Ilg2017Flownet2Evolution,Ranjan2017OpticalFlowEstimation,Jiang2021LearningEstimateHidden} were recently shown to be susceptible to adversarially modified input images, which alter the resulting attacked flow $\fadv$ to resemble a specified target flow $\targ$.
The few existing adversarial attacks on optical flow methods generate either perturbations with small L$_p$ norms~\cite{Schrodi2022TowardsUnderstandingAdversarial,Schmalfuss2022PerturbationConstrainedAdversarial} or adversarial patches~\cite{Ranjan2019AttackingOpticalFlow}, while adversarial weather attacks are completely unexplored.
In contrast, adversarial perturbations that imitate snow effects have been investigated in the context of classification~\cite{Kang2019TestingRobustnessUnforeseen,Marchisio2022FakeweatherAdversarialAttacks} or human pose estimation~\cite{Wang2021WhenHumanPose}.
However, for these applications, weather attacks~\cite{Kang2019TestingRobustnessUnforeseen,Marchisio2022FakeweatherAdversarialAttacks} or snow augmentations~\cite{Hendrycks2018BenchmarkingNeuralNetwork,Michaelis2019BenchmarkingRobustnessObject} only have to be applied to single images rather than sequences.
For optical flow estimation, a realistic motion of the weather effect over multiple frames and camera perspectives is required, imposing certain geometric constraints in time, which prevents the direct application of existing single-image adversarial weather generation schemes.
Also, learning models for variations in images from data rather than modeling them explicitly has been explored for adversarial training of robust classification methods~\cite{Gowal2020AchievingRobustnessWild,Robey2020ModelBasedRobust,Wong2020LearningPerturbationSets} or to synthesize snowy versions of satellite images~\cite{Ren2020DeepSnowSynthesizing}.
To ensure a realistic 3D motion of the weather effect in time, our attack explicitly models the motion of snow particles.
This enables the computation of a ground-truth optical flow field for scenes with generated snow as the motion of each snowflake is known, which would not directly be possible with learned weather models.
In terms of visual quality, the results of previous snow attacks were so far moderately convincing~\cite{Kang2019TestingRobustnessUnforeseen,Marchisio2022FakeweatherAdversarialAttacks} compared to conventional, non-differentiable rendering of snow effects~\cite{Bernuth2019SimulatingPhotoRealistic}.

\medskip \noindent
\textbf{Contributions.}
(i)~In our paper, we present a differentiable snow-to-scene rendering framework that generates visually appealing snow, which moves realistically over multiple times steps.
(ii)~Based on this rendering framework, we devise the first adversarial snow attack for optical flow. It optimizes 3D spatial positions of snowflakes in the scene rather than 2D per-pixel perturbations, resulting in attacked images that retain high realism in snow movement and appearance.
(iii)~And finally, our snow attack not only leads to a significant degradation of optical flow results, but also illustrates that methods with little sensitivity to small L$_p$ perturbations are particularly affected when the snow-parameters are optimized.

\section{Adversarial snow}
\label{sec:advsnow}

\begin{figure}
\centering
\begin{tikzpicture}
\draw (0, 0) node[inner sep=0,anchor=north west] (img) {
\includegraphics[width=.8\linewidth]{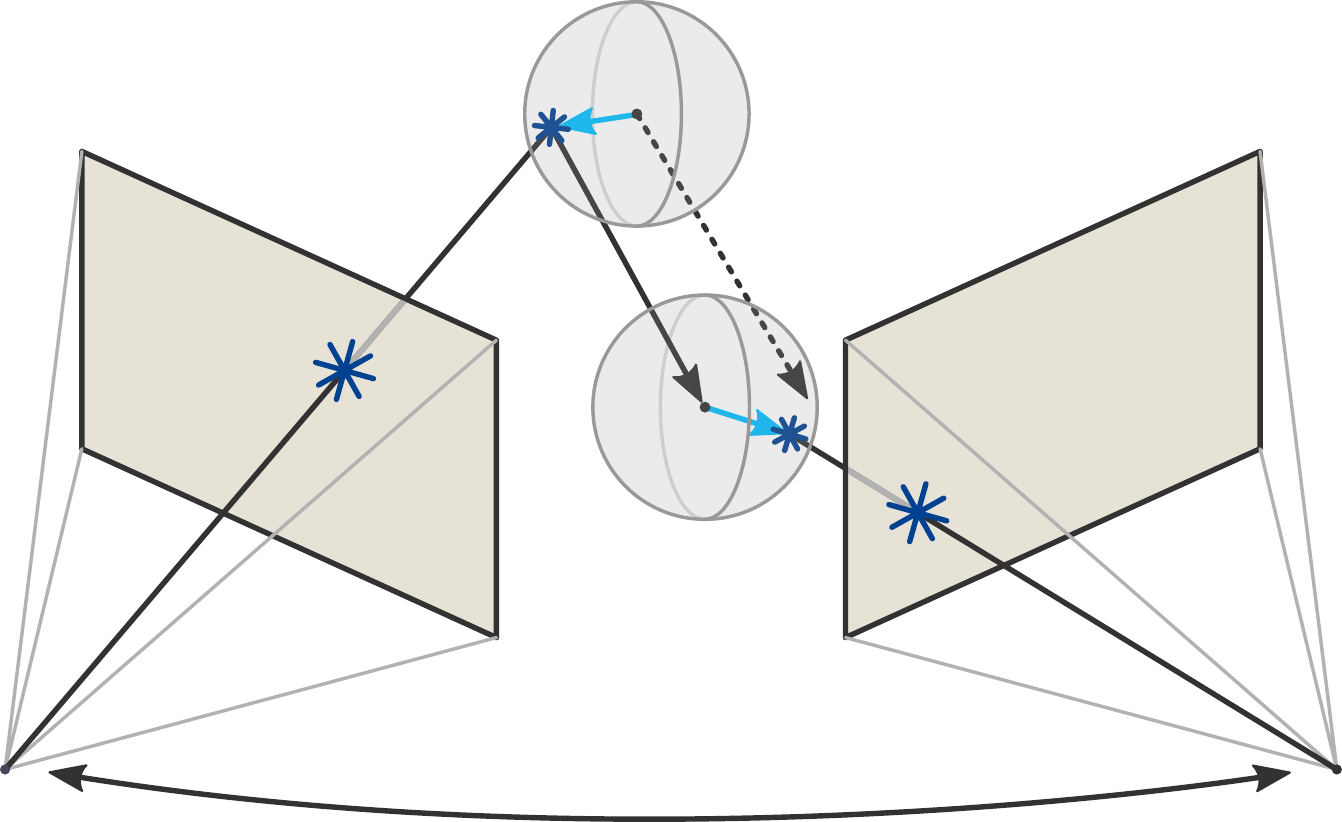}
};
\draw (3, -0.4) node {\footnotesize $\delta_t$};
\draw (3.7, -2.3) node {\footnotesize $\delta_{t\!+\!1}$};
\draw (3.4, -3.9) node {\footnotesize $A$};
\draw (2.2, -1) node {\footnotesize $d_t$};
\draw (5.4, -3.26) node {\footnotesize $d_{t\!+\!1}$};
\end{tikzpicture}
\caption{Model for snowflake motion in the 3D space.}
\label{fig:pointsvis}
\end{figure}
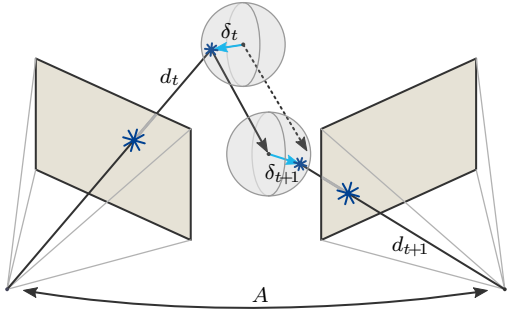

To study the robustness of optical flow algorithms towards weather effects, we design an adversarial attack that augments image sequences with snowfall.
To this end, we first equip an image sequence with parametrized snowflakes of realistic appearance and motion.
Second, we optimize the snowflake parameters such that the resulting snowy image sequence causes a specified (wrong) flow prediction.

This approach imposes three constraints on the creation of the snowflakes:
(i)~Because motion estimation is designed to cope with moving objects in a 3D scene, a simple 2D animation of the snowflakes in the image plane is not sufficient.
Instead, we have to model a realistic 3D motion, which also respects camera motion and object depth.
(ii)~Moreover, expanding our pursuit of realism also to the appearance of the snow, the snowflakes should be integrated with appropriate visual effects.
Such effects include an occlusion-aware depth placement as well as an out-of-focus blur.
(iii)~Finally, to enable the adversarial optimization of the snowflake, the whole rendering of the parametrized snowflakes needs to be differentiable.

\subsection{Snow generation and rendering}

To create 2D images of spatio-temporally consistent and visually appealing snowflakes we proceed in two steps.
First, we initialize a fixed set of snowflakes in the 3D scene and equip them with properties: Initial 3D positions, 3D motion, 3D offsets before and after the motion ($\de$, $\dt$), shapes, scaling and transparencies $\transp$ (see Fig.~\ref{fig:pointsvis} for the motion model).
Second, we make use of the 3D scene information to differentiably render the snowflakes in both frames, assuming that depth and camera information is given.
\begin{figure}
\centering
\setlength{\fboxrule}{0.1pt}%
\setlength{\fboxsep}{0pt}%
\begin{tabular}{@{}M{41mm}@{\ }M{41mm}@{}}
    \darkleftrightbox{41mm}{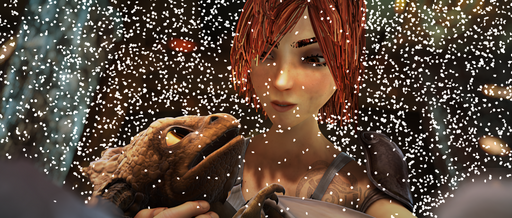}{Flake initialization}{Frame $t\vphantom{+}$} & \darkleftrightbox{41mm}{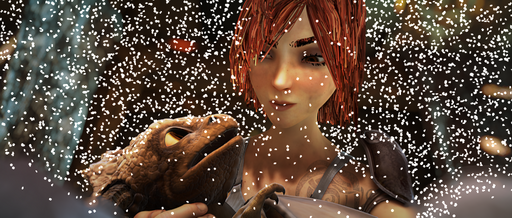}{}{Frame $t\!+\!1$} \\[-1pt]
    \darkleftbox{41mm}{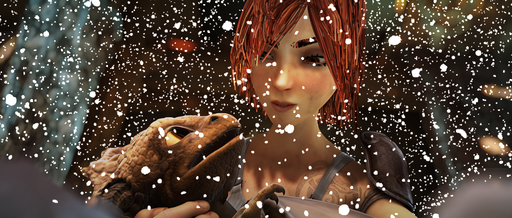}{+ Scaling} & \includegraphics[width=41mm]{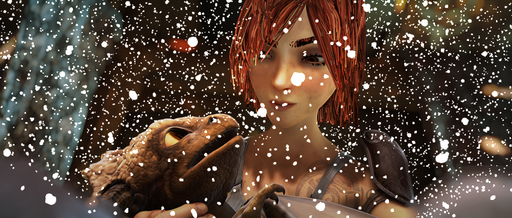}{} \\[-1pt]
    \darkleftbox{41mm}{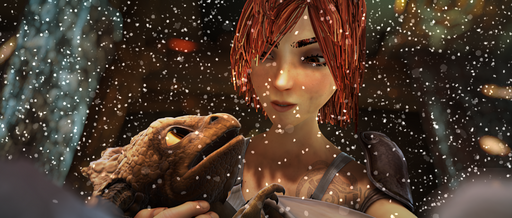}{+ Transparency} & \includegraphics[width=41mm]{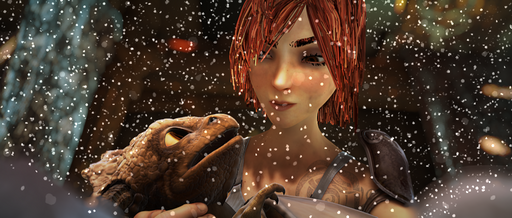}{} \\[-1pt]
    \darkleftbox{41mm}{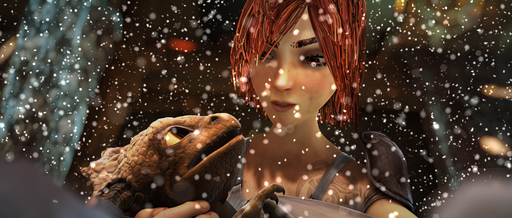}{+ Blur} & \includegraphics[width=41mm]{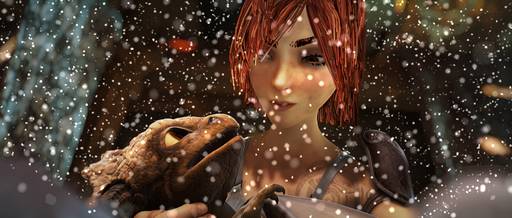}{} \\[-1pt]
    \darkleftbox{41mm}{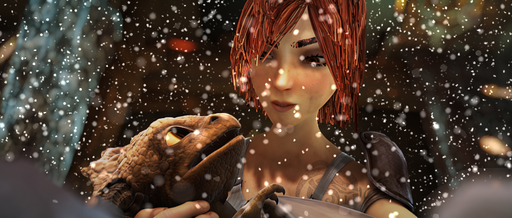}{+ Occlusion-aware} & \includegraphics[width=41mm]{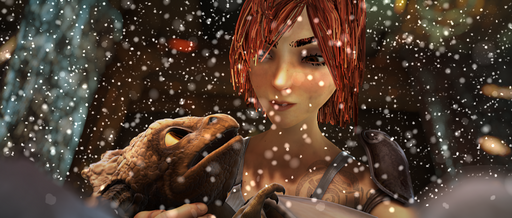}{} \\
\end{tabular}
\caption{Breakdown of our realistic snow rendering process.}
\label{fig:snowAugmentation}
\end{figure}

\medskip \noindent
\textbf{Snowflake initialization.}
To initialize the snowflake positions, we uniformly sample a fixed number of points in the 3D scene that are visible in the first frame or -- after adding the 3D motion -- in the second frame.
Every snowflake is associated with a 2D flake image, which is randomly sampled from a set of snowflake templates and rotated by a random angle (Fig.~\ref{fig:snowAugmentation}, row 1).
Then, we scale the flake image according to the snowflake's inverse depth, and initialize the transparency with a depth-dependent value (rows 2, 3).
Finally, we add realistic out-of-focus blur by convolving the flake image with a point spread function (row 4).

\medskip \noindent
\textbf{Snowflake rendering.}
We render the snowflakes with their associated 3D positions and transparencies in the given input frames as follows:
First, we compute the corresponding 2D points in both frames, which yields center positions for the 2D flake images.
Using the camera projec\-tion matrix and the relative transformation matrix $A$, we project the 3D points and their motion-displaced positions into the first and second frame, respectively.
Next, we determine the visibility for each pixel of the flake-image by interpolating a visibility map computed from frame depth and the flake depth $d$ per camera, which allows a realistic, occlusion-aware scene integration (Fig.~\ref{fig:snowAugmentation}, last row).
Lastly, we add the 2D flake images at the correct subpixel locations to the frames through bilinear interpolation, which enables a differentiation w.r.t.\ the snowflake parameters.

\subsection{Adversarial optimization}

After the snowflakes $\mathcal{S}$ are initialized and rendered, we modify certain snowflake parameters to change the output $\fadv$ of optical flow networks towards a desired target flow $\targ$.
We optimize transparency $\transp$ and offsets $\de$, $\dt$ before and after the motion.
Other parameters like initial 3D positions, 3D motion and 2D flake image are fixed.
To ensure a valid range of transparency values, we transform the bounded variable to a continuous one before optimization.
Our loss function measures flow differences with the average endpoint error (AEE)~\cite{Schmalfuss2022PerturbationConstrainedAdversarial} and allows larger offsets $\de$, $\dt$ for distant snowflakes via an $\alpha$-balanced MSE-like term, weighted with the inverse of the snowflake depth $d$:
\begin{equation}
\loss(\fadv, \targ, \mathcal{S}) = \text{AEE}(\fadv, \targ) +\! \sum_{i  \in t,t\!+\!1} \frac{\alpha_i}{|\mathcal{S}|} \sum_{s\in\mathcal{S}} \frac{\|\delta_i(s)\|_2^2}{d_i(s)}
\end{equation}

\section{Experiments}
\label{sec:experiments}

We implement the snow attack in PyTorch and generate snowy versions for all sequences of the Sintel dataset~\cite{Butler2012NaturalisticOpenSource}, as it is the de-facto standard for optical flow benchmarking and provides depth and camera information.
Attacked are the optical flow methods GMA~\cite{Jiang2021LearningEstimateHidden}, FlowNet2~\cite{Ilg2017Flownet2Evolution}, SpyNet~\cite{Ranjan2017OpticalFlowEstimation} with Sintel checkpoints, which were identified in~\cite{Schmalfuss2022PerturbationConstrainedAdversarial} to represent approaches with either high quality / low robustness, medium quality and robustness or low quality / high robustness, respectively.
We choose a zero-flow target $\targ$ (white flow visualization)~\cite{Schmalfuss2022PerturbationConstrainedAdversarial}, offset penalty weights $\alpha_t\!=\!\alpha_{t\!+\!1}\!=\!1000$ and snow falling down:right (ratio 5:2).
In several experiments, we investigate the attack strength AEE$(\fadv,\targ)$ (small value when attacked flow and target coincide), when (i) the number of snowflakes increases, (ii) different snow parameters are optimized, and (iii) optimized snow for one method is transferred to another.

\medskip \noindent
\textbf{Snow density.}
First, we optimize $\de$, $\dt$ and $\transp$, and study the attack strength when the total number of snowflakes in both sequences is varied from 1000 to 5000. 
In Table~\ref{table:numflakes}, the attack strength (distance from attacked flow to target) of adversarial snow increases with the snowflake number, independent of the attacked method.
On average, the distance is more than halved from 1000 to 5000 snowflakes.
This demonstrates the realistic behavior of our adversarial snow, which increases its influence with the covered image area.
We use 3000 snowflakes for further experiments because they balance attack strength and visual snow density.

\begin{table}
\begin{center}
\begin{tabular}{@{\ }lrrrrr@{\ }}
\toprule
Method & 1000 & 2000 & 3000 & 4000 & 5000 \\
\midrule
SpyNet   & 7.04 & 5.44 & 4.54 & 4.14 & \textbf{3.62} \\
FlowNet2 & 6.33 & 4.33 & 3.19 & 2.38 & \textbf{2.28} \\
GMA      & 8.56 & 6.38 & 4.42 & 3.38 & \textbf{2.29} \\
\bottomrule
\end{tabular}
\caption{Attack strength AEE$(\fadv,\targ)$ of adversarial snow with an increasing \emph{number of snowflakes} per frame-pair (1000--5000) on different optical \emph{flow methods}, best attack strength is bold.}
\label{table:numflakes}
\end{center}
\end{table}

\begin{table}
\begin{center}
\begin{tabular}{@{\ }lrrr@{\ }}
\toprule
Parameters & SpyNet & FlowNet2 & GMA\\
\midrule
Initial snow         & 13.29 & 21.93 & 12.25 \\
\midrule
$\de$             &  5.26 &  4.51 &  5.76 \\
$\dt$             &  6.53 &  4.72 &  6.88 \\
$\transp$               & 12.74 & 19.72 & 11.98 \\
\midrule
$\de$, $\dt$      &  \underline{4.68} &  \underline{3.73} &  \textbf{4.41} \\
$\de$, $\transp$        &  5.15 &  4.11 &  5.95 \\
$\dt$, $\transp$        &  6.21 &  4.84 &  6.89 \\
\midrule
$\de$, $\dt$, $\transp$ &  \textbf{4.54} &  \textbf{3.19} &  \underline{4.42}\\
\bottomrule
\end{tabular}
\caption{Attack strength AEE$(\fadv,\targ)$ of adversarial snow, optimized for combinations of \emph{snow parameters} $\de$, $\dt$ and $\transp$. \emph{Initial snow} measures the attack strength of randomly initialized snow.}
\label{table:optimvariables}
\end{center}
\end{table}

\begin{figure*}
\centering
\setlength{\fboxrule}{0.1pt}%
\setlength{\fboxsep}{0pt}%
\begin{tabular}{@{}M{19.1mm}@{}M{19.1mm}@{}M{19.1mm}@{\ }M{19.1mm}@{}M{19.1mm}@{}M{19.1mm}@{\ }M{19.1mm}@{}M{19.1mm}@{}M{19.1mm}@{}}
    \darkleftbox{19.1mm}{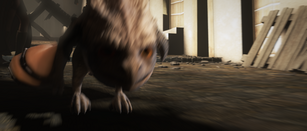}{\scriptsize Original} &
    \includegraphics[width=19.1mm]{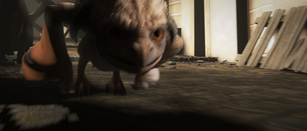} &
    \fcolorbox{gray!50}{white}{\includegraphics[width=19.1mm]{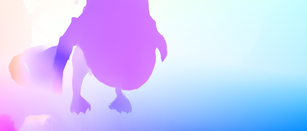}} &
    \darkleftbox{19.1mm}{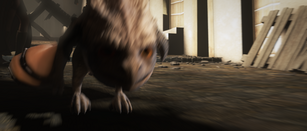}{\scriptsize Original} &
    \includegraphics[width=19.1mm]{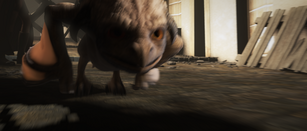} &
    \fcolorbox{gray!50}{white}{\includegraphics[width=19.1mm]{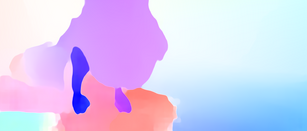}} &
    \darkleftbox{19.1mm}{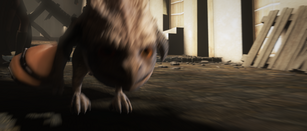}{\scriptsize Original} &
    \includegraphics[width=19.1mm]{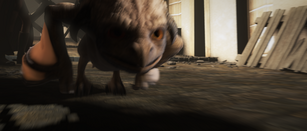} &
    \fcolorbox{gray!50}{white}{\includegraphics[width=19.1mm]{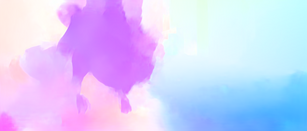}}
    \\[-3.5pt]
    \darkleftbox{19.1mm}{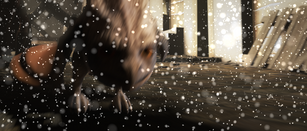}{\scriptsize Random} &
    \includegraphics[width=19.1mm]{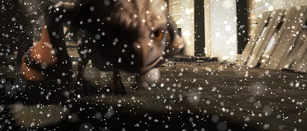} &
    \fcolorbox{gray!50}{white}{\includegraphics[width=19.1mm]{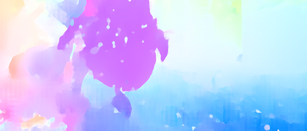}} &
    \darkleftbox{19.1mm}{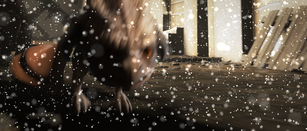}{\scriptsize Random} &
    \includegraphics[width=19.1mm]{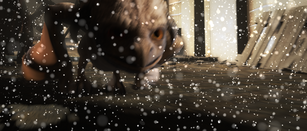} &
    \fcolorbox{gray!50}{white}{\includegraphics[width=19.1mm]{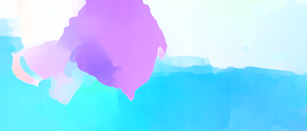}} &
    \darkleftbox{19.1mm}{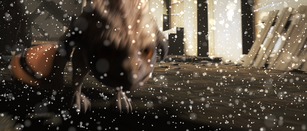}{\scriptsize Random} &
    \includegraphics[width=19.1mm]{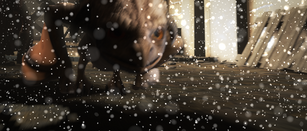} &
    \fcolorbox{gray!50}{white}{\includegraphics[width=19.1mm]{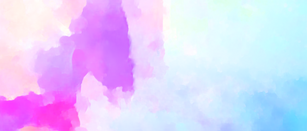}}
    \\[-3.5pt]
    \darkleftbox{19.1mm}{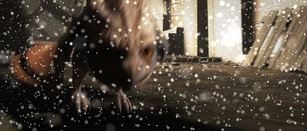}{\scriptsize Adversarial} &
    \includegraphics[width=19.1mm]{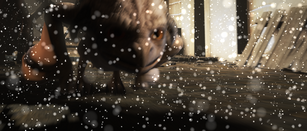} &
    \fcolorbox{gray!50}{white}{\includegraphics[width=19.1mm]{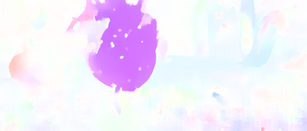}} &
    \darkleftbox{19.1mm}{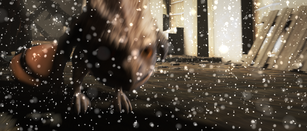}{\scriptsize Adversarial} &
    \includegraphics[width=19.1mm]{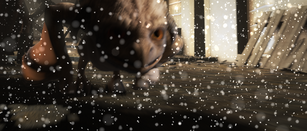} &
    \fcolorbox{gray!50}{white}{\includegraphics[width=19.1mm]{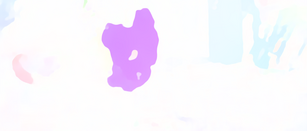}} &
    \darkleftbox{19.1mm}{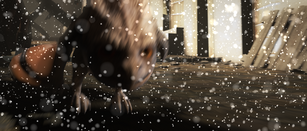}{\scriptsize Adversarial} &
    \includegraphics[width=19.1mm]{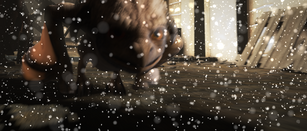} &
    \fcolorbox{gray!50}{white}{\includegraphics[width=19.1mm]{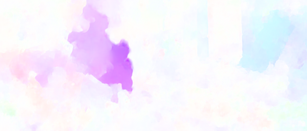}}
    \\
    \darkleftbox{19.1mm}{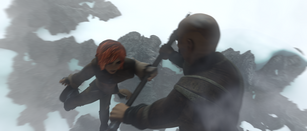}{\scriptsize Original} &
    \includegraphics[width=19.1mm]{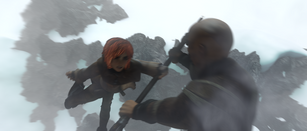} &
    \fcolorbox{gray!50}{white}{\includegraphics[width=19.1mm]{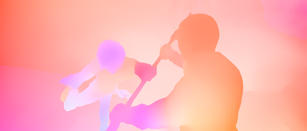}} &
    \darkleftbox{19.1mm}{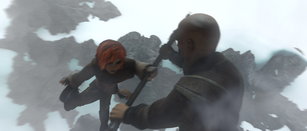}{\scriptsize Original} &
    \includegraphics[width=19.1mm]{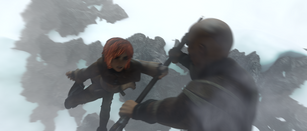} &
    \fcolorbox{gray!50}{white}{\includegraphics[width=19.1mm]{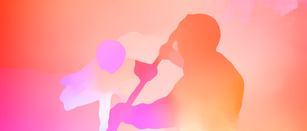}} &
    \darkleftbox{19.1mm}{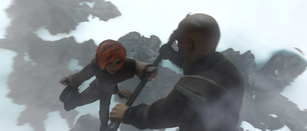}{\scriptsize Original} &
    \includegraphics[width=19.1mm]{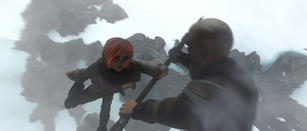} &
    \fcolorbox{gray!50}{white}{\includegraphics[width=19.1mm]{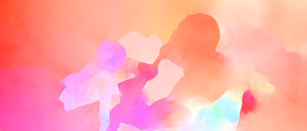}}
    \\[-3.5pt]
    \darkleftbox{19.1mm}{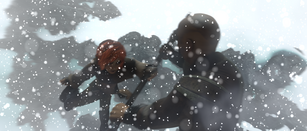}{\scriptsize Random} &
    \includegraphics[width=19.1mm]{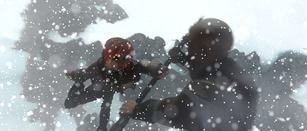} &
    \fcolorbox{gray!50}{white}{\includegraphics[width=19.1mm]{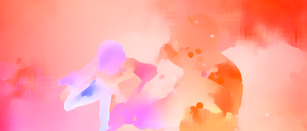}} &
    \darkleftbox{19.1mm}{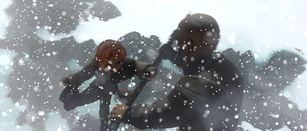}{\scriptsize Random} &
    \includegraphics[width=19.1mm]{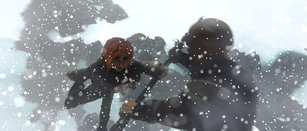} &
    \fcolorbox{gray!50}{white}{\includegraphics[width=19.1mm]{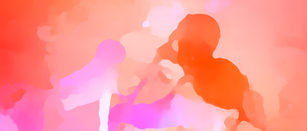}} &
    \darkleftbox{19.1mm}{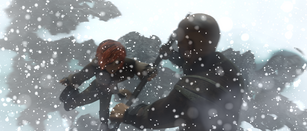}{\scriptsize Random} &
    \includegraphics[width=19.1mm]{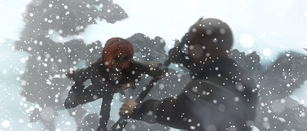} &
    \fcolorbox{gray!50}{white}{\includegraphics[width=19.1mm]{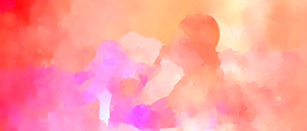}}
    \\[-3.5pt]
    \darkleftbox{19.1mm}{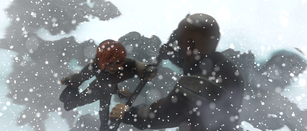}{\scriptsize Adversarial} &
    \includegraphics[width=19.1mm]{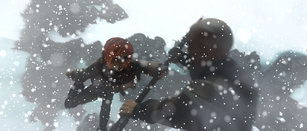} &
    \fcolorbox{gray!50}{white}{\includegraphics[width=19.1mm]{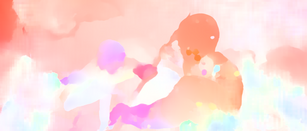}} &
    \darkleftbox{19.1mm}{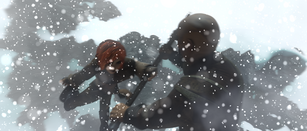}{\scriptsize Adversarial} &
    \includegraphics[width=19.1mm]{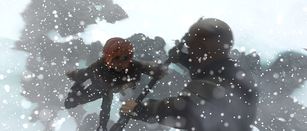} &
    \fcolorbox{gray!50}{white}{\includegraphics[width=19.1mm]{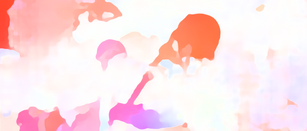}} &
    \darkleftbox{19.1mm}{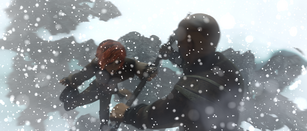}{\scriptsize Adversarial} &
    \includegraphics[width=19.1mm]{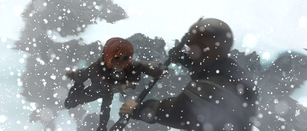} &
    \fcolorbox{gray!50}{white}{\includegraphics[width=19.1mm]{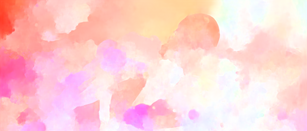}}
    \\
    \darkleftbox{19.1mm}{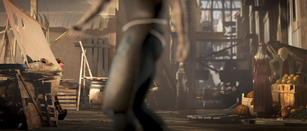}{\scriptsize Original} &
    \includegraphics[width=19.1mm]{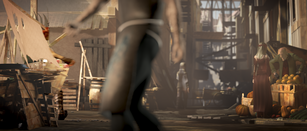} &
    \fcolorbox{gray!50}{white}{\includegraphics[width=19.1mm]{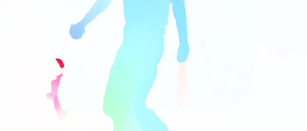}} &
    \darkleftbox{19.1mm}{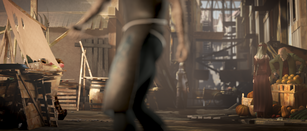}{\scriptsize Original} &
    \includegraphics[width=19.1mm]{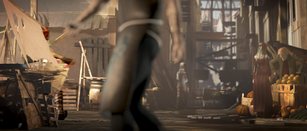} &
    \fcolorbox{gray!50}{white}{\includegraphics[width=19.1mm]{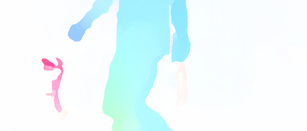}} &
    \darkleftbox{19.1mm}{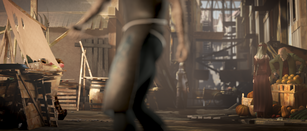}{\scriptsize Original} &
    \includegraphics[width=19.1mm]{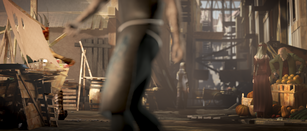} &
    \fcolorbox{gray!50}{white}{\includegraphics[width=19.1mm]{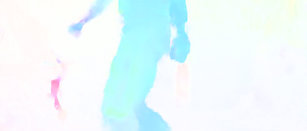}}
    \\[-3.5pt]
    \darkleftbox{19.1mm}{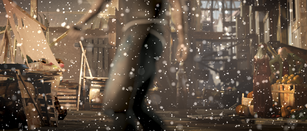}{\scriptsize Random} &
    \includegraphics[width=19.1mm]{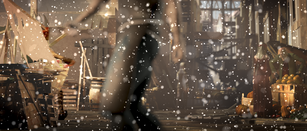} &
    \fcolorbox{gray!50}{white}{\includegraphics[width=19.1mm]{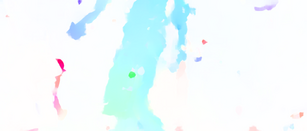}} &
    \darkleftbox{19.1mm}{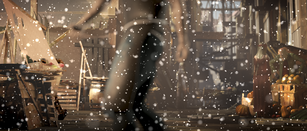}{\scriptsize Random} &
    \includegraphics[width=19.1mm]{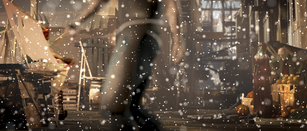} &
    \fcolorbox{gray!50}{white}{\includegraphics[width=19.1mm]{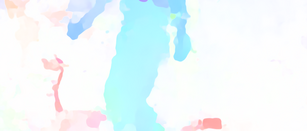}} &
    \darkleftbox{19.1mm}{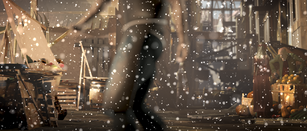}{\scriptsize Random} &
    \includegraphics[width=19.1mm]{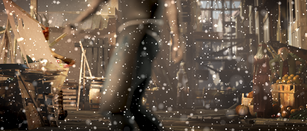} &
    \fcolorbox{gray!50}{white}{\includegraphics[width=19.1mm]{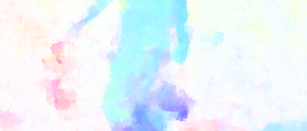}}
    \\[-3.5pt]
    \darkleftbox{19.1mm}{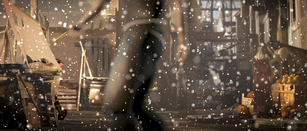}{\scriptsize Adversarial} &
    \includegraphics[width=19.1mm]{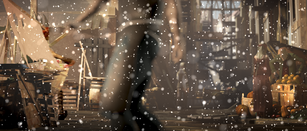} &
    \fcolorbox{gray!50}{white}{\includegraphics[width=19.1mm]{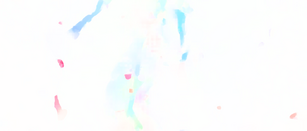}} &
    \darkleftbox{19.1mm}{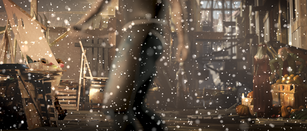}{\scriptsize Adversarial} &
    \includegraphics[width=19.1mm]{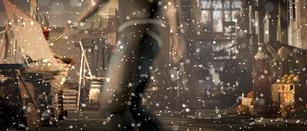} &
    \fcolorbox{gray!50}{white}{\includegraphics[width=19.1mm]{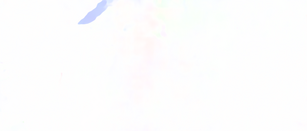}} &
    \darkleftbox{19.1mm}{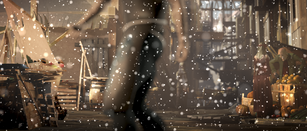}{\scriptsize Adversarial} &
    \includegraphics[width=19.1mm]{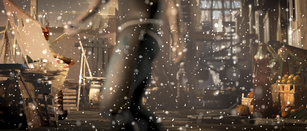} &
    \fcolorbox{gray!50}{white}{\includegraphics[width=19.1mm]{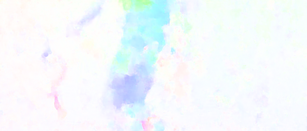}}
    \\
\end{tabular}
\caption{Qualitative results for 3000 snowflakes on images from the Sintel final dataset with \emph{random} initialization and \emph{adversarial} optimization with optical flow predictions for GMA~\cite{Jiang2021LearningEstimateHidden}, FlowNet2~\cite{Ilg2017Flownet2Evolution} and SpyNet~\cite{Ranjan2017OpticalFlowEstimation} (\emph{left to right}).}
\label{fig:strongattack}
\end{figure*}

\medskip \noindent
\textbf{Optimizing snow attack strength.}
Next, we investigate how optimizing different snow parameters influences the attack strength.
Tab.~\ref{table:optimvariables} summarizes the attack strength for snow attacks with 3000 snowflakes, optimized for all parameter combinations, and compared to a random snow initialization.
In all combinations, optimizing the position offset in the first frame $\de$ yields the strongest attacks, while varying the snowflake transparency $\transp$ has the smallest impact.
Nonetheless, all parameters $\de$, $\dt$, $\transp$ must be optimized to reach the strongest attack - only for GMA it suffices to optimize $\de$ and $\dt$ alone.
Fig.~\ref{fig:strongattack} visualizes the attacked frames and predicted flows for all methods when optimizing $\de$, $\dt$ and $\transp$.
There, we make several interesting observations:
When comparing randomly initialized and adversarially optimized snowflakes, their positions differ only slightly and the adversarial sample is indistinguishable from random snow to a human observer.
Also, it is surprising that snowflakes eradicate the estimated motion despite their inability to stand still due to falling and camera motion.
Moreover, when comparing the initial snow to the optimized results, FlowNet2 and SpyNet that were previously identified to be relatively robust~\cite{Schmalfuss2022PerturbationConstrainedAdversarial}, alter their predictions significantly in the presence of adversarial snow.
We ascribe this phenomenon to the more detailed flow estimations of GMA, which detects the localized motion of single snowflakes (\cf Fig.~\ref{fig:strongattack}, col.\ 3, where circular snowflakes are visible).
The less accurate methods FlowNet2 and SpyNet instead propagate the detected motion from snowflakes over larger areas, rather than attributing it to small moving objects (\cf Fig.~\ref{fig:strongattack}, col.\ 6,9, where optical flow predictions have few details).

\medskip \noindent
\textbf{Snow transferability.}
Finally, we investigate if snowflakes optimized for one method can change the flow predictions of another.
We use the adversarial snow attack with 3000 snowflakes and optimize the snow for $\de$, $\dt$ and $\transp$, before evaluating all networks on the resulting adversarial images.
Tab~\ref{table:transfer} summarizes the results.
While snowflakes are most effective on the method they were optimized for, transferred snow has a measurable negative impact on FlowNet2 and GMA compared to random snow (\cf Tab.~\ref{table:optimvariables}, Init).
For even more transferable configurations, snowflakes could be optimized over several images and methods.

\begin{table}[tb]
\small
\begin{center}
\begin{tabular}{l|@{\ \ \ }ccc}
\toprule
Test
\begin{tikzpicture}[overlay]
\draw (-0.32,0.28) -- (1.22,-0.11);
\end{tikzpicture}
\hspace{8mm}
Train & SpyNet & FlowNet2 & GMA \\
\midrule
  SpyNet & \cellcolor{gray!20}\textbf{\phantom{0}4.54} &        13.80  &        13.44  \\
FlowNet2 &        16.54  & \cellcolor{gray!20}\textbf{\phantom{0}3.19} &        14.21  \\
     GMA &        10.33  &        10.67  & \cellcolor{gray!20}\textbf{\phantom{0}4.42} \\
\bottomrule
\end{tabular}
\caption{Transferability of adversarial snow strength AEE$(\fadv,\targ)$ of snowflake positions that were optimized for one \emph{optical flow method} to another, best attack in bold.}
\label{table:transfer}
\end{center}
\end{table}

\section{Conclusion}
\label{sec:conclusion}

In this paper we proposed a novel adversarial attack on motion estimation algorithms with realistic snow.
To this end, we developed a differentiable snowflake renderer that can be used to generate adversarial samples with a strong impact on optical flow methods.
Interestingly, our attack demonstrates the ability to let networks predict zero-flow although the snowflakes undergo both individual and camera motion.
At the same time, the resulting attacked images are visually indistinguishable from random snow images, making our attack unnoticeable for a human observer.
Finally, more accurate methods appear to be more robust towards adversarially optimized snow than towards small L$_p$ perturbations, as they detect the motion of single snowflakes rather than propagating the motion into the wider image.

\medskip
\noindent
\textbf{Acknowledgments.}
Funded by the Deutsche For\-schungs\-gemeinschaft (DFG, German Research Foundation) -- Project-ID 251654672 -- TRR 161 (B04).
Jenny Schmalfuss is supported by the International Max Planck Research School for Intelligent Systems (IMPRS-IS).

{\small
\bibliographystyle{ieee_fullname}
\bibliography{egbib}
}

\end{document}